\documentclass[conference]{IEEEtran}
\IEEEoverridecommandlockouts
\usepackage{cite}
\usepackage{amsmath,amssymb,amsfonts}
\usepackage{algorithmic}
\usepackage{graphicx}
\usepackage{textcomp}
\usepackage{xcolor}
\usepackage{svg}
\usepackage[outdir=./]{epstopdf}
\usepackage{calc}
\usepackage{enumerate} 
\usepackage{enumitem} 
\usepackage{multicol}
\def\BibTeX{{\rm B\kern-.05em{\sc i\kern-.025em b}\kern-.08em
    T\kern-.1667em\lower.7ex\hbox{E}\kern-.125emX}}
\begin{document}

\title{A Scalable Predictive Maintenance Model for Detecting Wind Turbine Component Failures Based on SCADA Data}

\author{\IEEEauthorblockN{Lorenzo Gigoni, Alessandro Betti}
\IEEEauthorblockA{\textit{i-EM S.r.l} \\
Livorno, Italy \\
lorenzo.gigoni@i-em.eu}

\and
\IEEEauthorblockN{Mauro Tucci, Emanuele Crisostomi}
\IEEEauthorblockA{\textit{University of Pisa} \\
Pisa, Italy \\
mauro.tucci@unipi.it}}

\maketitle

\begin{abstract}
In this work, a novel predictive maintenance system is presented and applied to the main components of wind turbines. The proposed model is based on machine learning and statistical process control tools applied to SCADA (Supervisory Control And Data Acquisition) data of critical components. The test campaign was divided into two stages: a first two years long offline test, and a second one year long real-time test. The offline test used historical faults from six wind farms located in Italy and Romania, corresponding to a total of 150 wind turbines and an overall installed nominal power of 283 MW. The results demonstrate outstanding capabilities of anomaly prediction up to 2 months before device unscheduled downtime. Furthermore, the real-time 12-months test confirms the ability of the proposed system to detect several anomalies, therefore allowing the operators to identify the root causes, and to schedule maintenance actions before reaching a catastrophic stage.
\end{abstract}


\section{Introduction}
Wind energy stands out nowadays as one of the most promising alternatives to conventional dispatchable energy sources. However, unscheduled downtime and components replacements represent an urgent issue to be addressed in order to mitigate the impact of operation and maintenance (O\&M) costs over generation, and maintain wind energy an attractive and competitive choice \cite{b1}. In this context, SCADA-based Condition Monitoring \cite{b2,b3} recently emerged as a promising solution to shift from an expensive reactive maintenance strategy to a pre-emptive, predictive one. Traditional approaches include bivariate analysis based on power curve modeling \cite{b4} and condition parameter-based models \cite{b5, b6}. The latter approach, roughly speaking, consist in training over healthy wind turbine (WT) data samples and predicting the status of the WT by monitoring residuals between the forecasted and the measured parameters. However, while condition parameter-based models give some information of specific components affecting the WT operation, both such traditional methods fail to provide a comprehensive picture of the correlations among the component parameters. 
This work aims to overcome the drawbacks of the aforementioned methodologies; namely, the main objective is to accurately predict anomalies for the three most critical WT components (gearbox, generator bearing and main bearing) within a multivariate framework, with the final goal of leading to more informed maintenance decisions. The model exploits signals related to the energy conversion system and specific components. The proposed approach provides many novel contributions: 
\begin{enumerate}[label=(\roman*)]
\item machine learning techniques and process control tools, i.e. the residual Hotelling T2 control chart \cite{b7,b8}, are wisely combined;
\item an innovative multivariate outliers removal (MOR) method, based on k-means clustering, is proposed and applied to eliminate the abnormal samples from training instances (the proposed MOR approach generalizes the procedures already presented in \cite{b4, b9});
\item the status of the components is monitored by means of an original probabilistic formula defining a Key Performance Indicator (KPI): warnings of different severity are triggered based on threshold crossing rules; 
\item the Plug and Play nature of the presented approach, i.e. the fast service scalability on wind farms of increasing size, is also an added value of this work;
\item the proposed systems was realised and tested on a large number of wind turbines in different farms and geographical areas, during three years of operation.
\end{enumerate}

\section{Model Description}

\begin{figure}[!b]
\centerline{\includegraphics[scale= .7]{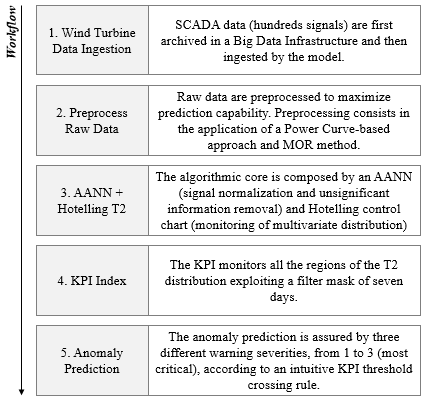}}
\caption{Predictive Maintenance Model Workflow. The model is composed by five main steps. First, SCADA data are imported (1) then the dataset is preprocessed to maximize the model prediction capability (2). Later signals are transformed in normally-distributed tags and monitored by the Hotelling T2 chart (3), from which a KPI is finally computed (4) to uncover predictive insights(5) }
\label{fig:model_workfllow}
\end{figure}

The algorithmic core of the model is represented by the combination of an Auto-Associative Neural Network (AANN), used for achieving normally distributed signals, and a variant of the multivariate Hotelling T2 control chart, which supervises a specific subset of SCADA tags for each WT component (Fig. \ref{fig:model_workfllow}). The model is fed with signals related to both the energy conversion system (e.g. active power, wind speed, shafts speed, shaft torque, etc.) and component-specific (e.g. gear box oil temperature, gear bearing temperature, etc.). The subset of component tags was selected from the list of available plants signals (originally of the order of hundreds) by combining specific domain knowledge, as well as feature selection strategies. 

The proposed multivariate control chart is built using historical data that correspond to healthy operation periods, according to on-site information. The historical dataset of multivariate instances undergoes several preprocessing operations, namely:
\begin{enumerate}[label=(\roman*)]
\item outliers are removed by means of a bivariate power curve-based approach;
\item data samples are clustered using  k-means clustering method, exploiting the classic Euclidean distance metric to group samples into clusters;
\item 	component temperatures are seasonally adjusted by taking advantage of a least-squares fitting against outdoor temperature in the low load operative regime. 
\end{enumerate}
This operation allows to extend the lead time before incipient faults are predicted and reduces the rate of false positives.

The pre-processed data is then used to train an AANN, which enforces an information compression on the hidden layer, and transforms data to be roughly normally distributed, which is known to be a desired feature for the subsequent T2 approach \cite{b9,b10,b11}. The multivariate output samples from the AANN are finally fed to the control chart, which detects changes in the underlying non-linear dynamics of the system by means of an original KPI formula. Such KPI, based on a probabilistic approach, monitors the population size housing in different T2 distribution sub-regions and penalizes deviations from the normal behavior, which corresponds to KPI values close to 1. Warning levels of different severity are finally triggered according to threshold crossing rules and shown on a Business Intelligence Analytics dashboard. The model workflow is shown in Fig.1.

\section{Results and Discussion}

\begin{figure}[!b]
\centerline{\includegraphics[scale= 0.29]{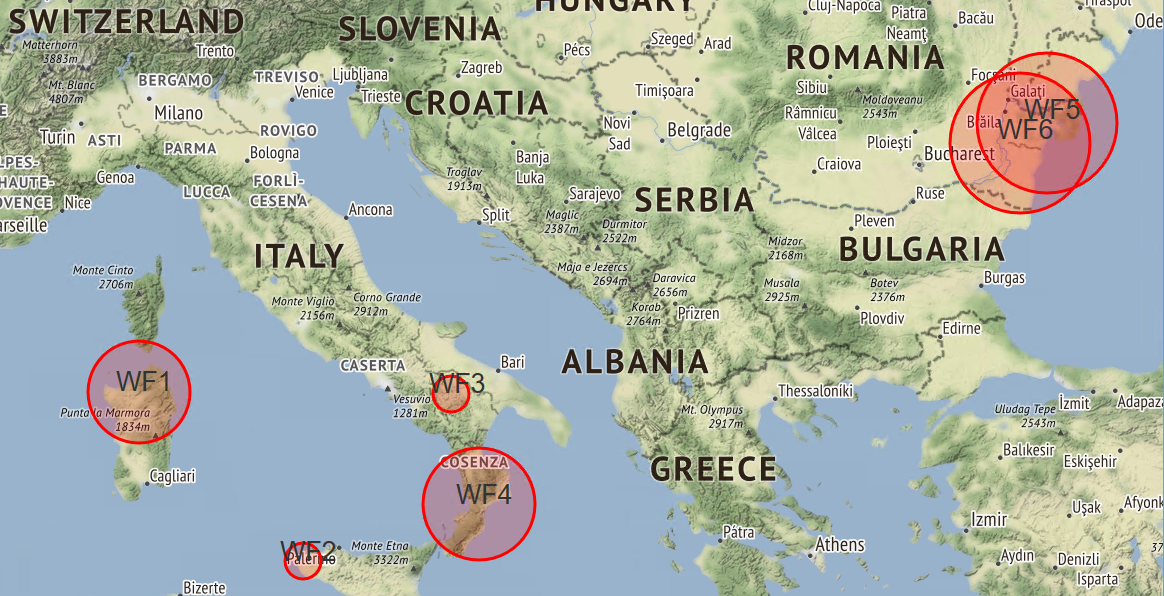}}
\caption{Position of the 6 WFs, of which 4 are located in Southern Italy and Sardegna, and the remaining 2 in Romania. Globally, the model was evaluated over150 wind turbines and an overall nominal power of 283 MW. In particular, the bubble size in figure is proportional to the wind farm size.}
\label{fig:windfarm_map}
\end{figure}

The model was tested on six wind farms located in Italy and Romania corresponding to four different WT manufacturers and two different sizes (1,5 MW and 2 MW), see Fig. \ref{fig:windfarm_map}. Globally, the model was evaluated over 150 wind turbines and 283 MW of nominal power. The positions of the wind farms are shown in Fig. 2. The large amount of SCADA data (Historical and Real-Time) have been archived and preliminary preprocessed by means of the Big Data infrastructure Microsoft Data Lake Analytics, before feeding the model. 

\begin{figure*}[!t]
\centering
\includegraphics[scale= 0.70]{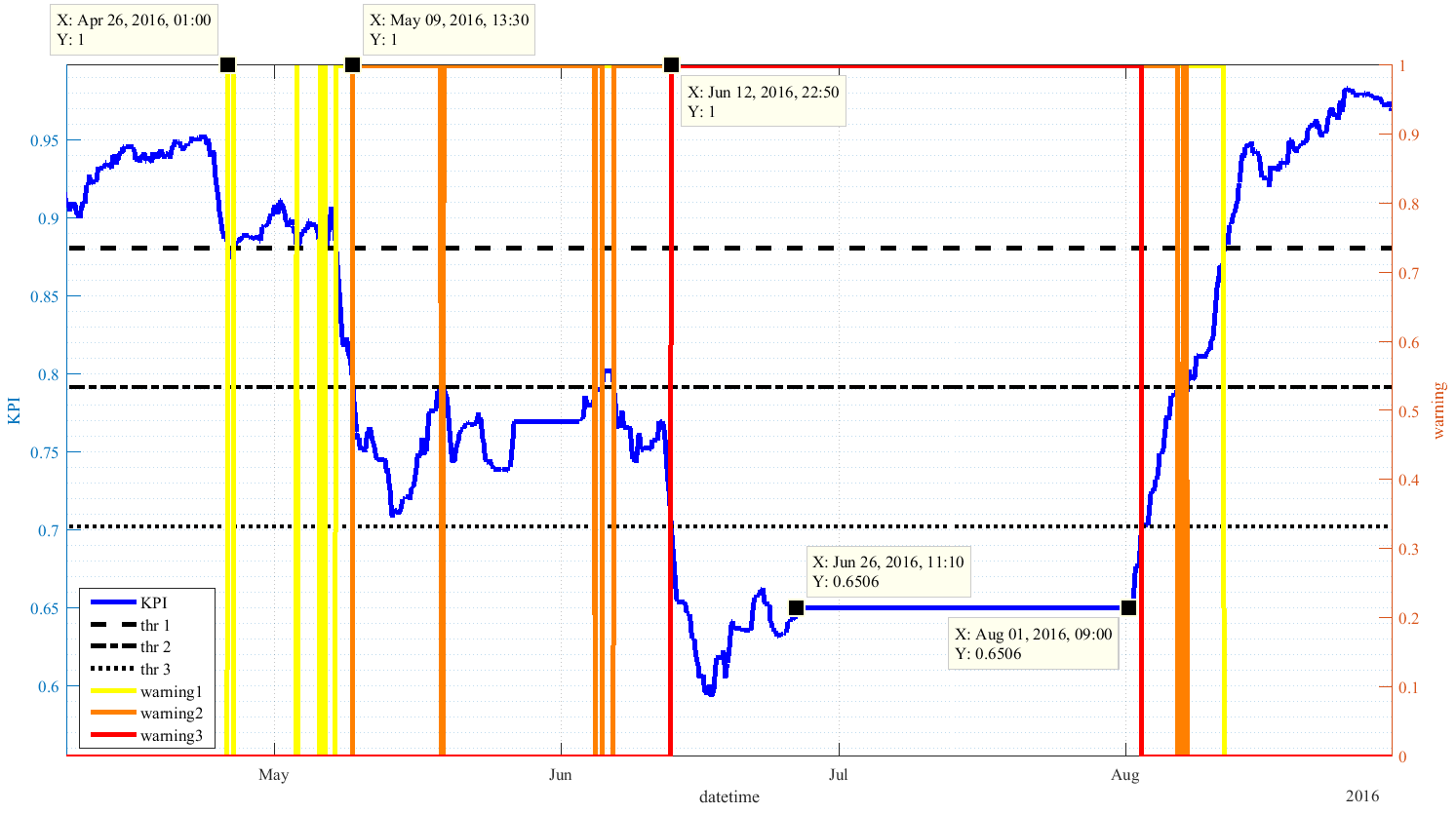}
\caption{Historical Validation: Supervision model output for the generator bearings of a Wind Turbine in Italy. On the left (right) the KPI (warnings of different severities) are shown as a function of time. The model early detected a generator bearing issue on 26th of April, i.e. at least 2 months before the operators detection. This problem involved a 1 month outage, which corresponds to the flat portion of the KPI in figure. In addition to the first warning level triggered on April, the model detected two further deviations from nominal condition: on the 9th of May it triggered warning level 2 and then on the 12th June it triggered the most severe warning (warning 3).}
\label{fig:offline_case}
\end{figure*} 

\begin{figure*}[t!]
\centering
\includegraphics[scale= 0.75]{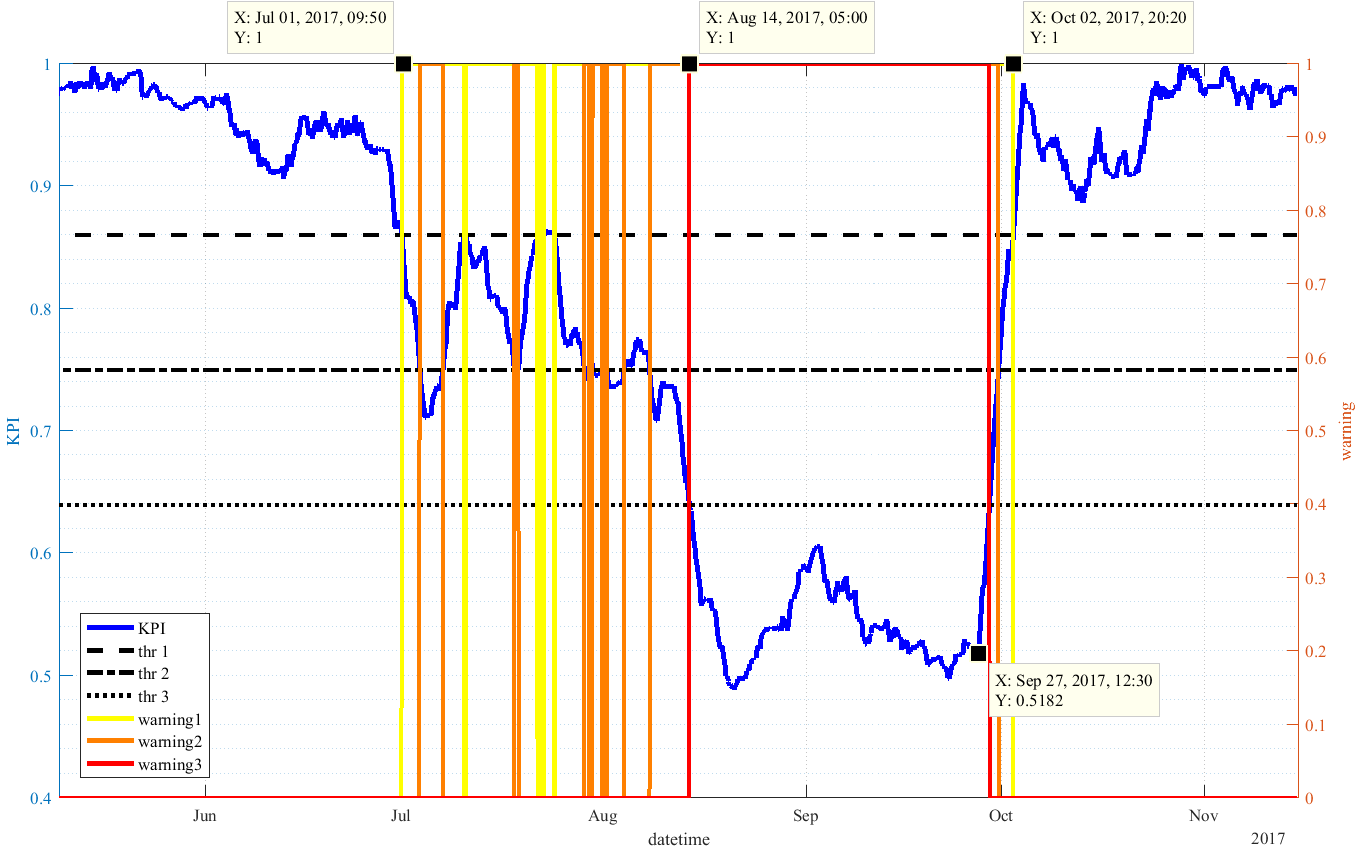}
\caption{Real-Time Phase: anomaly occurring at the gearbox of a WT in Romania. On the left (right) the KPI (warnings) are shown as a function of time. The model detected on the 1st of July an anomaly at the gearbox. The operators, thanks to the service provided, identified the root cause (gearbox cooling system issue) and scheduled maintenance activities avoiding further problem to the gearbox.}
\label{fig:online_case}
\end{figure*}

\begin{figure*}[t!]
\centering
\includegraphics[scale= 0.73]{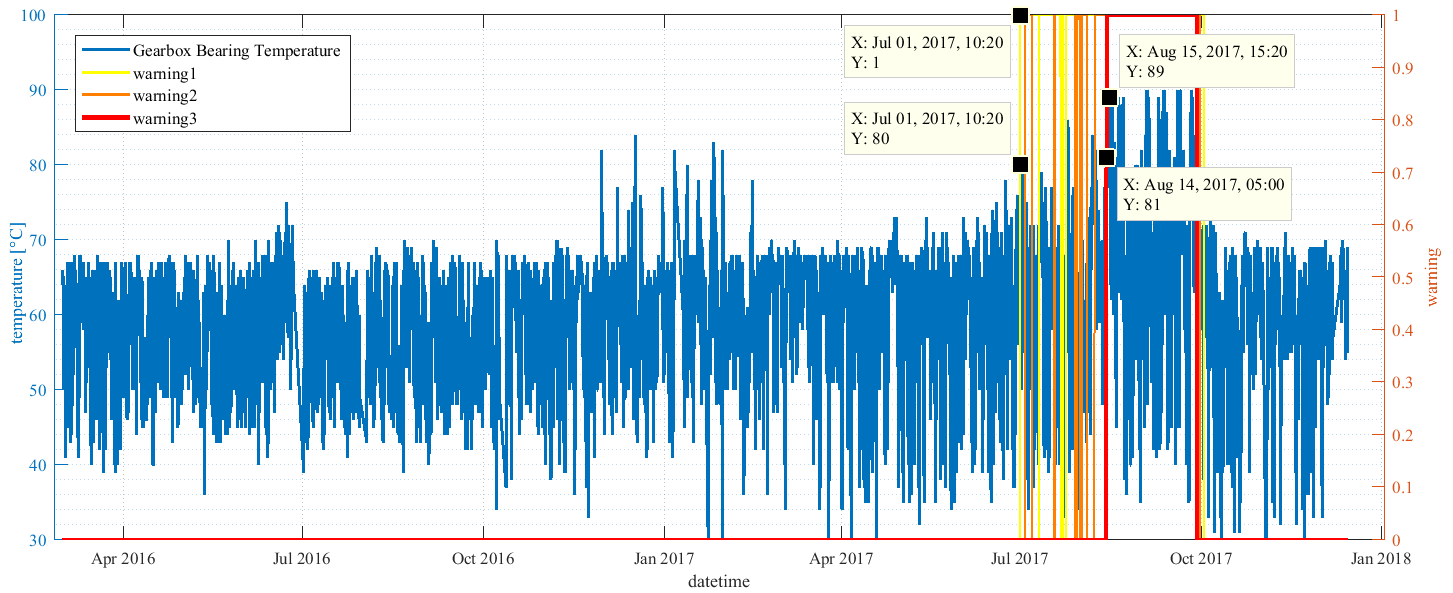}
\caption{Real-Time Phase: anomaly occurring at the gearbox of a WT in Romania. On the left (right) the gearbox bearing temperature (warnings) are shown as a function of time. On the 1st of July 2017 the model detected an anomaly at the gearbox, where the gearbox bearing temperature reached values over 80 °C, rarely shown during the previous year. Later, the model triggered the most severe warning on the 14th of August, corresponding to a peak in the gearbox bearing temperature of almost 90 °C.}
\label{fig:online_diagnostic}
\end{figure*}

\subsection{Historical Validation}

\begin{table}[!b]
  \centering
  \caption{Historical Cases exploited for model tuning and validation.}
    \begin{tabular}{ccccc}

    \multicolumn{4}{c}{\textbf{Wind Farm Characteristics}} & \multicolumn{1}{c}{\textbf{Anomalous Cases }}\\
    \#    &\textit{ No. WT} & \textit{WT Power} & \textit{Manufacturer} & \textbf{Delivered } \\
\hline
    WF1   & 34    & 1.5   & A     & - \\

\hline
    WF2   & 9     & 2.0   & B     & - \\
\hline
    WF3   & 9     & 2.0   & B     & - \\
\hline
    WF4   & 28    & 2.0   & B     & 3 Gearbox\\
& & & & 
5 Generator\\
\hline
    WF5   & 35    & 2.0   & C     & 1 Generator \\

\hline
    WF6   & 35    & 2.0   & D     & 3 Generator \\
    \end{tabular}%
  \label{tab:offline_cases}%
\end{table}%

The software was trained and validated on historical data (2015-2016) using case studies (Tab. \ref{tab:offline_cases}) provided by plant operators. The obtained results highlighted an outstanding prediction strength of bearing failures from 2 weeks up to 1-2 months ahead, in good agreement with other research papers \cite{b2, b6}. 

Fig. \ref{fig:offline_case}  shows the output of the supervision model for the generator bearing of a Wind Turbine in Italy. On the left the KPI is shown as a function of time (blue curve), whereas on the right warnings of different severities (yellow, orange and red in ascending order of severity) are drawn, corresponding to different KPI threshold crossings. This clearly demonstrates the predictive power of the proposed KPI as explained below.

\begin{table}[!b]
  \centering
  \caption{Anomalous cases delivered during Real-Time stage.}
    \begin{tabular}{ccccc}

    \multicolumn{4}{c}{\textbf{Wind Farm Characteristics}} & \multicolumn{1}{c}{\textbf{Anomalous Cases }}\\
    \#    &\textit{ No. WT} & \textit{WT Power} & \textit{Manufacturer} & \textbf{Delivered } \\
\hline
    WF1   & 34    & 1.5   & A     & 2 Gearbox \\

& & & &
4 Generator \\
\hline
    WF2   & 9     & 2.0   & B     & 1 Generator \\
\hline
    WF3   & 9     & 2.0   & B     & No Cases Delivered \\
\hline
    WF4   & 28    & 2.0   & B     & 2 Generator \\
& & & & 
2 Main Bearings \\
\hline
    WF5   & 35    & 2.0   & C     & 3 Gearbox \\
& & & &
1 Generator \\
\hline
    WF6   & 35    & 2.0   & D     & 10 Gearbox \\
    \end{tabular}%
  \label{tab:delivered_cases}%
\end{table}%

In particular, the aforementioned WT suffered a generator bearing fault during the year 2016. At that time no predictive strategy was applied, and the issue was detected by plant operators on the 26th of June 2016 and involved a device outage which lasted for 36 days (up to 1st August), as shown by the flat portion of the blue curve in Fig. \ref{fig:offline_case}.
The ex-post application of supervision model anticipates the plant operator detection of about two months. Indeed the model triggers a first level warning on the 26th of April, where KPI falls down crossing warning level 1 threshold (dashed line). After fifteen days, the model detects a further deviation from the nominal behavior of the generator bearing, which corresponds to the KPI crossing the warning level 2 threshold (dash-dot line) on the 9th of May. Finally, two weeks before the outage (i.e. 12th June) the model triggers the most severe warning (warning 3), corresponding to the most critical device condition. After maintenance intervention on the generator bearing , the KPI comes back to nominal condition and the model restarts to monitor the status of the device. Table I summarizes the characteristics of the six wind farms (number of WTs and their nominal power) and shows the anomalies detected during the offline campaign.  
The offline validation campaign has therefore demonstrated outstanding capabilities of early stage detection of WT main components anomalies, which basically corresponds, in the aforementioned case, to a prediction of generator bearing faults. 

\subsection{Real-Time Phase}

Downstream of the the model tuning and validation campaign, the predictive service was installed on-site and tested on the year 2017, in order to verify its compatibility with plant operator activities. As a result, the model delivered 25 anomalous cases during one year of real time operation (Tab. \ref{tab:delivered_cases}). 
 
Of these, we received a response from local plant operators for almost one half, and more than 90\% of the latter corresponded really to true positives. The predicted were failures mainly related to gearbox, and in a smaller part to generator and main bearings (15, 8, and 2, respectively). Wind farm operators, exploiting model warnings, were able to apply efficient Predictive Maintenance strategies anticipating on-site actions from 2 weeks up to 1-2 months with respect to their traditional O\&M activities. The capability of the model to reveal different fault classes was confirmed thanks to feedbacks received from local plant operators. In fact, the model also revealed cooling system issues, such as dirt accumulation,and the breaking of the engine fan, affecting generator and gearbox bearings. Furthermore, the model was capable of early detecting incipient heat dissipation issues after grease replacement in the generator bearings. 

Fig. \ref{fig:online_case} shows  an example of an anomaly detected by the model and occurring at a gearbox component of a WT located in Romania. During July 2017, the WT suffered an issue at the gearbox cooling system. In particular, the dirt accumulation involved an inefficient heat dissipation. On the 1st July the model early detected an abnormal gearbox temperature and triggered the first level warning to the wind farm operators. The operators benefited from the real-time service, receiving continuous feedback of the anomaly by monitoring the KPI time series, which show further deviations after the 1st of July, up to triggering the most severe warning on the 14th of August. Finally, they identified the root cause of the  anomaly and scheduled the maintenance activities at the end of September, avoiding any further problems to the gearbox, which restored its full operation after the 2nd of October (Fig. \ref{fig:online_case}).
A qualitative analysis of the model output for the case described above is shown in  Fig. \ref{fig:online_diagnostic}, where the gearbox bearing temperature (blue curve, left axis) and warnings (right axis) are shown as a function of time. In correspondence of warning level 1, the gearbox bearing temperature reaches over 80 °C, rarely seen during the previous year. Later, when the model triggered the most severe warning on the 14th of August, the gearbox bearing arrived at abnormal temperatures of 90 °C or higher, well above the nominal operational temperatures.

The Real-Time service confirmed therefore the validity of the proposed approach and its contribution to extend the component lifetimes, as well as to move from a traditional reactive or time-based maintenance activity towards a predictive maintenance strategy.

\section{conclusion}
 In this paper a novel predictive maintenance model for WT sub-components failures was presented, based on the integration of Machine Learning and statistical process control tools, and applying a multivariate approach at all the processing levels. The model is fed with SCADA tags and is composed mainly by four processing phases: (i) an outliers removal phase based on power curve modeling and k-means clustering, (ii) a seasonally adjusting procedure for temperature signals, (iii) an AANN-based residual approach to normalize tags, (iv) and finally a residual control chart based on a novel KPI formula to trigger components warnings. Results from a vast campaign on 150 wind turbines demonstrate the ability of the proposed methodology to predict failures at gearbox, generator and main bearing levels, with lead times up to 1-2 months, and ensuring early detection at least.

The availability of a large amount of data from different spatial, temporal and technological points of view, and the fast service deployment on new wind farms represent, at the best of our knowledge, a distinct feature of the proposed system with respect to other published studies.

%


\end{document}